\documentclass[letterpaper, 10 pt, conference]{ieeeconf}  

\IEEEoverridecommandlockouts                              
\usepackage{graphics} 
\usepackage{epsfig} 
\usepackage{mathptmx} 
\usepackage{times} 
\usepackage{amsmath} 
\usepackage{amssymb}  
\usepackage{fancyhdr}
\numberwithin{equation}{section}		

\title{\LARGE \bf
Learning Obstacle Avoidance using Double DQN for Quadcopter Navigation
}

\author{Nishant Doshi$^{1}$, Amey Sutavani$^{2}$ and Sanket Gujar$^{3}$
\thanks{*This work was not supported by arxic
any organization}
\thanks{$^{1}$Nishant Doshi is with the Department of Robotics Engineering, Worcester Polytechnic Institute, MA, 01609, USA {\tt\small ndoshi@wpi.edu}}%
\thanks{$^{2}$Amey Sutavani is with the Department of Robotics Engineering, Worcester Polytechnic Institute, MA, 01609, USA {\tt\small aasutavani@wpi.edu}}%
\thanks{$^{3}$Sanket Gujar is with the Department of Computer Science, Worcester Polytechnic Institute, MA, 01609, USA {\tt\small srgujar@wpi.edu}}%
}

\begin{document}

\maketitle
\thispagestyle{empty}
\pagestyle{empty}

\begin{abstract}
One of the challenges faced by Autonomous Aerial
Vehicles is reliable navigation through urban environments.
Factors like reduction in precision of Global Positioning System
(GPS), narrow spaces and dynamically moving obstacles make the
path planning of an aerial robot a complicated task. One of the
skills required for the agent to effectively navigate through such an environment is to develop an ability to avoid collisions using information from onboard depth sensors. In this paper, we propose Reinforcement Learning of a virtual quadcopter robot agent equipped with a Depth Camera to navigate through a simulated urban environment.
\end{abstract}



\section{INTRODUCTION}

In recent years, Quadcopters have been extensively used
for civilian task like object tracking, disaster rescue, wildlife
protection and asset localization. It presents interesting
application avenues especially in tasks such as automated mail
delivery system, fire protection and disaster management.
However, quadcopter navigation through urban environments
is a complex task because of frequent dynamic obstacles
(Humans, Posters, etc.). Also, the GPS navigation system can
perform poorly when surrounded by tall buildings in urban
environment, dilating the precision of the 3D position fix. It
becomes more dangerous when the quadcopter is flying
through tight spaces and is uncertain of its position, increasing
the chances of collision. The quadcopter also needs to take
smart action after detecting dynamic obstacles (Humans,
Vehicles, animals, traffic signals etc.) during navigation in
runtime in urban environment. Traditionally, obstacle
avoidance techniques have been designed as end point solution
in an aerial robot navigation. One of the promising approach
for this problem is deep reinforcement learning. In this paper a
simple model is developed for the task of detecting and
avoiding common civilian obstacles encountered by a
quadcopter while navigating a path in an urban environment.

From the reinforcement learning view, the main challenge
here is that, the policy should update itself during runtime for
stochastic obstacles detected in the environment and take the
optimal action accordingly. Also, the navigation problem has
sparse distributed reward in state space which is a challenge for
learning the shortest distance.

The objective of this project is to train a quadcopter to
navigate without hitting obstacles and taking a shortest path
around through a high-rise urban environment where stochastic
and dynamic obstacles are frequent.

The organization of the paper is as follows: Section I
provides a general introduction to the challenges for
quadcopter urban navigation. Section II provides a
prerequisites required to understand the experiments. Section
III defines the problem outlining the agent used and the
environment. Section IV gives a brief description about the
AirSim simulator, while section V defines the solution
approaches for the problem defined. Section VI describes the
experiments and training and testing arena used. Section VII
discusses the results for the experiments, while section VIII
describes the future attempt that can be made and section IX
describes the challenges faced during the experiments.

\section{PREREQUISITES}

\subsection{Quadcopter control}
A quadcopter is a simple aerial vehicle comprised of a rigid
square frame with four rotors at the vertices of the frame. Each
of the four rotors is controlled by a single motor which controls
the rpm of the rotor and essentially the lift that the particular
rotor generates. The spin of the rotors is chosen such that the
diagonally opposite rotors spin in same direction and the
adjacent rotors spin in opposite direction to each other. Thus,
all the four rotors contributes to two inputs: 
\begin{enumerate}
\item Lift force: By the virtue of the thrust generated by the propellers on the rotors
\item Torque: by the virtue of the spin of the rotors. Hence, in
a hover state, the total lift generated by all four rotors is equal
to the weight of the quadcopter and the net torque exerted on
the quadcopter frame by all four rotors is zero (because of the
spin of the adjacent rotors).
\end{enumerate}
The position (x, y, z) and orientation (roll, pitch, yaw) of
the quadcopter can thus be controlled by just varying the rpm
of each of its rotors. For example, if the quadcopter has to move
forward, it has to pitch down by reducing the rpm of its front
rotors while increasing the rpm of its hind rotors. It is worth
noting that a quadcopter’s yaw angle is independent of its flight
velocity. Thus, the quadcopter can face in any direction while
executing any lateral velocity.

\subsection{Q-Learning}
Q-Learning is a value-based learning algorithm for
reinforcement learning. It is an off policy learning method
where the state-action values are iteratively bettered by
applying discounted Bellman equation [6].
An Optimal Policy is one by following which, the agent can
maximize its long running reward. Q learning updates the
action values using temporal error. The update step can be
expressed as:

$$
Q(s, a) = Q(s, a)+ \alpha(Q_{e}(s, a) - Q(s, a)){ ......(1)}
$$

where,

$$
Q_{e}(s, a) = R(s) + \gamma \max_{a}(Q(s', a)){ ......(2)}
$$
Where,

$\alpha$ is the Learning Rate

$\gamma$ is the Discount Factor

s is the Current State

a is the Action taken from the current state

s$^{\prime}$ is the state following the action a from state s

and R(s) being the immediate reward for state s.

As Q Learning is based on temporal error, it has a high bias
and also faces the problem of overestimation as estimation and
update are simultaneously carried out on a single function
mapping. Hence, enhanced approaches based on Q learning
like Double Q learning are often preferred.

\subsection{Off-Policy vs On-policy Reinforcement Learning}
Reinforcement learning algorithms can be generally
characterized as off-policy where they employ a separate target
behavior policy that is independent of action policy being
improved upon. The benefit of this separation is that the target
behavior policy will be more stable by sampling all actions,
whereas the action estimation policy can be greedy, thus
reducing the bias. Q learning is an off-policy algorithm as it
updates the Q values without making any assumptions about
the actual policy followed.
In contrast, On-policy directly uses the policy that is being
estimated to sample trajectories during training.

\subsection{Model Free Algorithms}
Model-free algorithms are used where there are high-
dimensional state and action spaces, where the transition
matrix is incredibly expensive to compute in space and time [4].
Model-free algorithms makes no effort in learning the
dynamics that governs how an agent interacts with the
environment. It directly estimates the optimal policy or optimal
value function by policy iterations or value iterations.
However, model-free algorithms needs a large number of
training examples for accurate policy approximations [7].

\subsection{Deep Q Learning}
Deep Q Learning uses Deep Neural Networks which take
the state space as input and output the estimated action value
for all the actions from the state. The target action value update
can be expressed as:
$$
Q(s, a) = R(s)+ \gamma \max _{a}(Q_{P}(s^\prime, a))
$$
Where,
$Q_{P}$ is the network predicted value for the state s$^{\prime}$.
After convergence, the optimal action can be obtained by
selecting the action value corresponding to the maximum Q
value.

An enhancement employed for better convergence of this
method is the use of experience buffer. This buffer records the
states, actions and associated rewards from the agent’s
experience and occasionally trains the Q network with this
buffer to retain the former experience. This buffer itself is
updated with the training epochs to keep the experience buffer
updated.

\subsection{Double DQN}

Deep Q Learning suffers from overestimation as it involves
choosing a maximum Q value which may contain non uniform
noise. This will slow down learning as the agent spends more
time exploring non optimal states. A solution to this problem
was proposed by Hado van Hasselt (2010) and called Double
Q-Learning [2]. In this algorithm, two Q functions are
independently learned: one function $(Q_{1})$ is used to determine
the maximizing action and second $(Q_{2})$ to estimate its value. 
Either $Q_{1}$ or $Q_{2}$ is randomly updated by:
$$
Q_{1}(s, a) = r + \gamma  Q_{2}(s, \max _{a}(Q_{1}(s^\prime, a))
$$

Or,

$$
Q_{2}(s, a) = r + \gamma  Q_{1}(s, \max _{a}(Q_{2}(s^\prime, a))
$$

\section{PROBLEM DEFINITION}

Quadcopter navigation through urban environments is a
complex task because of frequent stochastic obstacles, and the
poor accuracy in GPS navigation system when surrounded by
tall buildings in urban environment due to precision dilation.
The problem is particularly dangerous when the quadcopter is 
navigating through tight spaces and it is uncertain of its
position, increasing the chances of collision. The quadcopter
also needs to take smart action to detect and avoid stochastic
obstacles like buildings, humans, vehicles, animals, traffic
signals etc. in real time while parallel running a navigation task [3][4].
The agent will be provided with a starting point and a goal
location, the agent will also be provided with inputs from the
front centered camera to take intelligent navigation decisions [5].
We need the agent to navigate the environment safely from
start point to the target without colliding into obstacles in the
path. Here the state space is continuous while the action space
is discrete.

\section{AIRSIM SIMULATOR} 
Airsim [1] is an open-source platform aiming to narrow the
gap between simulations and reality in order to aid
development of autonomous vehicles. It is built on Unreal
Engine that offers physically and visually realistic simulation
for collecting a large amount of annotated data in a variety of
conditions and environments. It includes a physics engine that
can operate at a high frequency for real-time-hardware-in-the-
loop (HITL) simulations with support for popular
communication protocols like MavLink [10].

Airsim also provides access to control the quadcopter in
computer vision mode, where the physics engine is disabled
and there is no flight controller active.

Airsim can be interfaced with opensource autopilot
hardware such as PX4 Autopilot [11] and Ardupilot Controller
[12]. This allows reinforcement learning algorithms to be
trained in simulation and validated against the realistic sensor
data in real world.

\subsection{Vision API and Camera choices}

Airsim provides 6 image type which are Scene, depth-
planner, depth-perspective, depth-vis, disparity-normalized,
segmentation and surface-normal. The camera ID 0 to 4
corresponds to center front, left front, right front, center
downward and center rear respectively. The image type and
camera can be easily configured using the vision API calls or
using the setting json files for capturing training images.

\subsection{Collision Detection}
Unreal engine offers a rich collision detection system
optimized for different classes of collision meshes. Airsim
receives the impact position, impact normal vector and
penetration depth for each collision that occurred during the
rendering interval. Airsim Physics engine uses this data to
compute the collision response with Coulomb friction to
modify both linear and angular kinematics.

The collision information can be obtained using
getCollisionInfo API. This call returns an object that has
information not only whether collision occurred but also
collision position, surface normal and penetration depth.

\begin{figure}[!h]
\centering
\includegraphics[width=8cm]{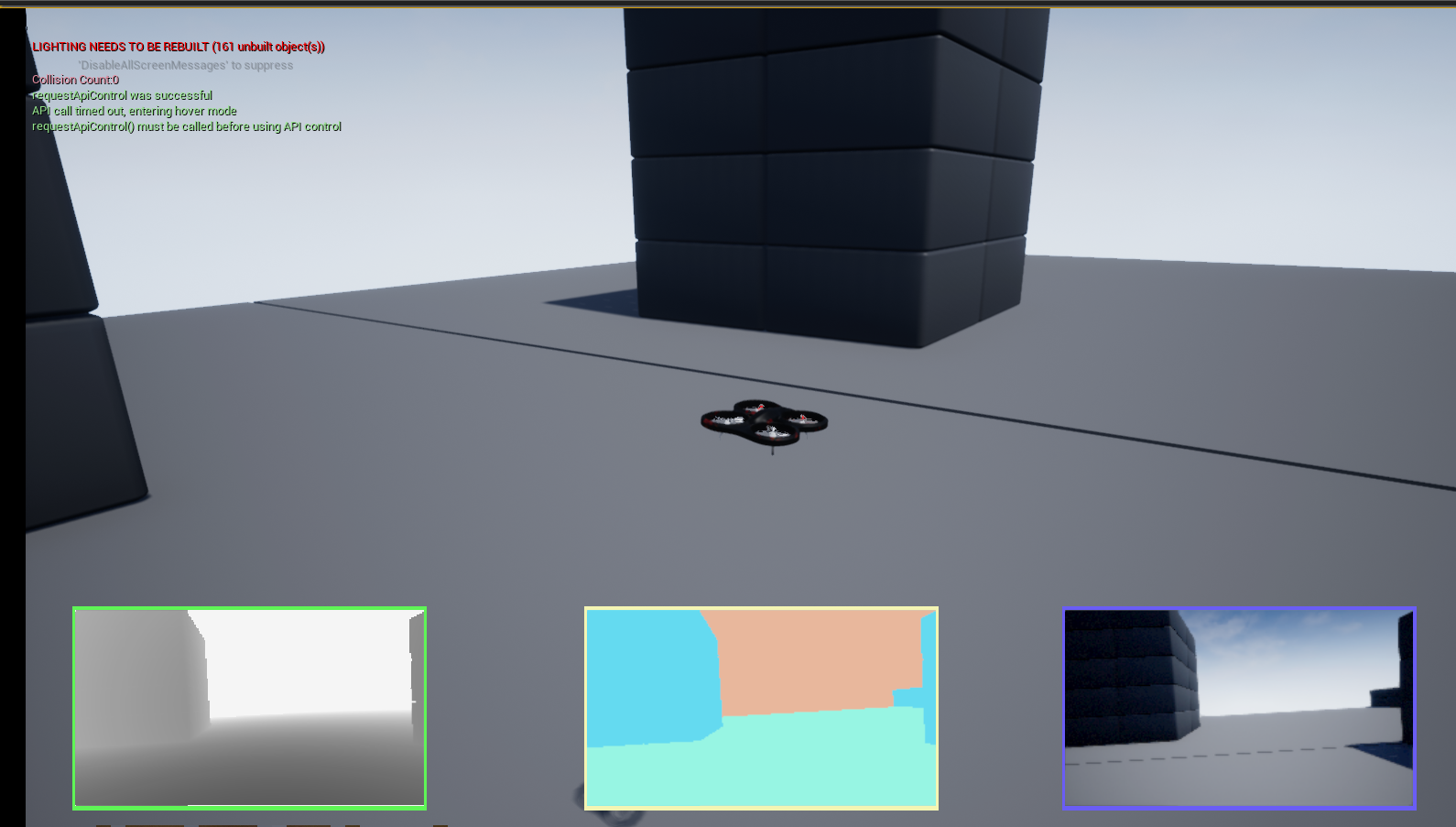}
\caption{AirSim Simulator View}
\end{figure}

\section{SOLUTIONS}

\subsection{Agent Description}
In every episode, our quadcopter agent will be “spawned”
in the simulated environment at the start point. The goal of our
quad agent is to reach the target location without colliding into
obstacles in the path. Here the state space is continuous while
the action space, comprising of the 5 yaw rates and a fixed
forward velocity, is discrete. We chose to implement DQN to
assist quadcopter to make intelligent decision for avoiding
obstacles and reaching the target location as quickly as
possible.

The quadcopter is equipped with a single front facing depth
camera where each pixel value corresponds to the actual depth
distance of the surroundings. We chose depth cameras over
standard RGB cameras to avoid artifacts due to lighting
conditions in the surroundings.

There were a few key challenges faced in controlling our
virtual quadcopter agent. We wanted the quadcopter to always
face the direction of its forward velocity to place the oncoming
obstacles on the front camera’s field of view. Therefore, it was
necessary to change the yaw angle of the quadcopter to the
angle of its velocity vector.

The physics engine employed by AirSim uses stochastic
flight controllers. Thus, all the actions commanded by our
learning algorithm were executed with a certain degree on
simulated noise in the simulator.

\subsection{Learning Architecture}
Since our state space is huge, it becomes imperative to use
function approximation to plausibly solve the task of collision
free navigation. Deep neural networks are good candidates for
this purpose. Furthermore, when coupled with Convolutional
Neural Networks, we can directly feed in camera images to the
networks to visually learn the navigation task [8].

Our DQN consist of 4 convolutional layers and 2 dense
layers with the output layer of the dimension as the action
space. DQN is feed a Depth Perspective image from the center
front camera of the quadcopter. We get depth image from
center camera by ray tracing each pixel. The resolution of the
depth image is 84x84.

The convolutional layers can be thought of as feature
extractors. The extracted features are then fed to the dense
layers which act as regression mechanism. The resulting
trained network maps a depth image to the corresponding
action to avoid collisions and while navigating to the goal.

The quadcopter is given a constant velocity in the forward
direction and there are 5 yaw-rate of (-10, -5, 0, 5, 10) degrees.

\subsection{Reward Function}
Reward is defined as a function of how far the quadcopter
is from the goal position and how far it is from the line joining
start and goal position. We consider the episode to terminate if
the quadcopter drifts too much from the start and goal position
joining vector, if it goes away from the goal beyond a fixed
threshold or it collides in the environment. We also constrain
the number of time steps which increases linearly with
episodes.


We observed that the quadcopter takes random actions in
the early episodes that sometimes makes it move in only small
area, so not exploring the complete environment as well as not
hitting any obstacles or moving towards/ away from the goal
position, so we decided to terminate the episodes if it reaches
the maximum action steps which increases linearly with
episode count.

\section{EXPERIMENTS}

In order for a safe autonomous flight the quadcopter
shouldn’t collide with any obstacles and should make
intelligent decisions, like changing route for avoiding
collisions. In this section, the experiments are arranged to
illustrate a quadcopter with the mission of reaching goal
position without colliding and taking minimum time. During
the flight the quadcopter constantly monitors the environment
with the depth-perspective image obtained from the center-
front camera.

The quadcopter also needs to reach the goal within a
defined action steps, so it also need to learn to optimize its path
to do so.

In order to represent the discussed scenarios, we came up
with two environments which are “Blocks” and custom
designed “Wobbles” training arena as shown in figure 4 and figure 5.

\subsection{Blocks Training arena}
The Blocks training arena is a rectangular shaped arena
with movable blocks spread across the arena. The blocks in the
environments can be moved and the arrangements can be
customized according to the requirements. The blocks arena
was designed to simulate simple construction structures like
buildings road squares.

Primitive Training: Initially, the agent was trained with no
goal position, so its reward was only dependent on collisions.
The approach was to teach the quadcopter to just avoid
obstacles. The agent starts at the initial position and is free to
explore the environment.

Testing and improving: The Initial position was kept fixed
at the center of the environment. The goal position were varied
to let the model to observe if it can avoid obstacles in any given
scenarios, like cutting the edge of obstacles in front, as well as
in side of the camera’s field of view. The first few episodes
were taken by the quadcopter to learn the direction towards the
goal position, the rest steps were taken to avoid the obstacles
encountered between the initial to goal state. The quadcopter
collided during some initial episodes but later learned deflect
itself from the edges. However, it was still sometimes
colliding, if it approaches the obstacles from the center where
there is not enough space for it to maneuver out of the
oncoming obstacle. This behavior is reflected in the results
where we can observe sudden drops in average rewards.

\begin{figure}[!h]
\centering
\includegraphics[width=8cm]{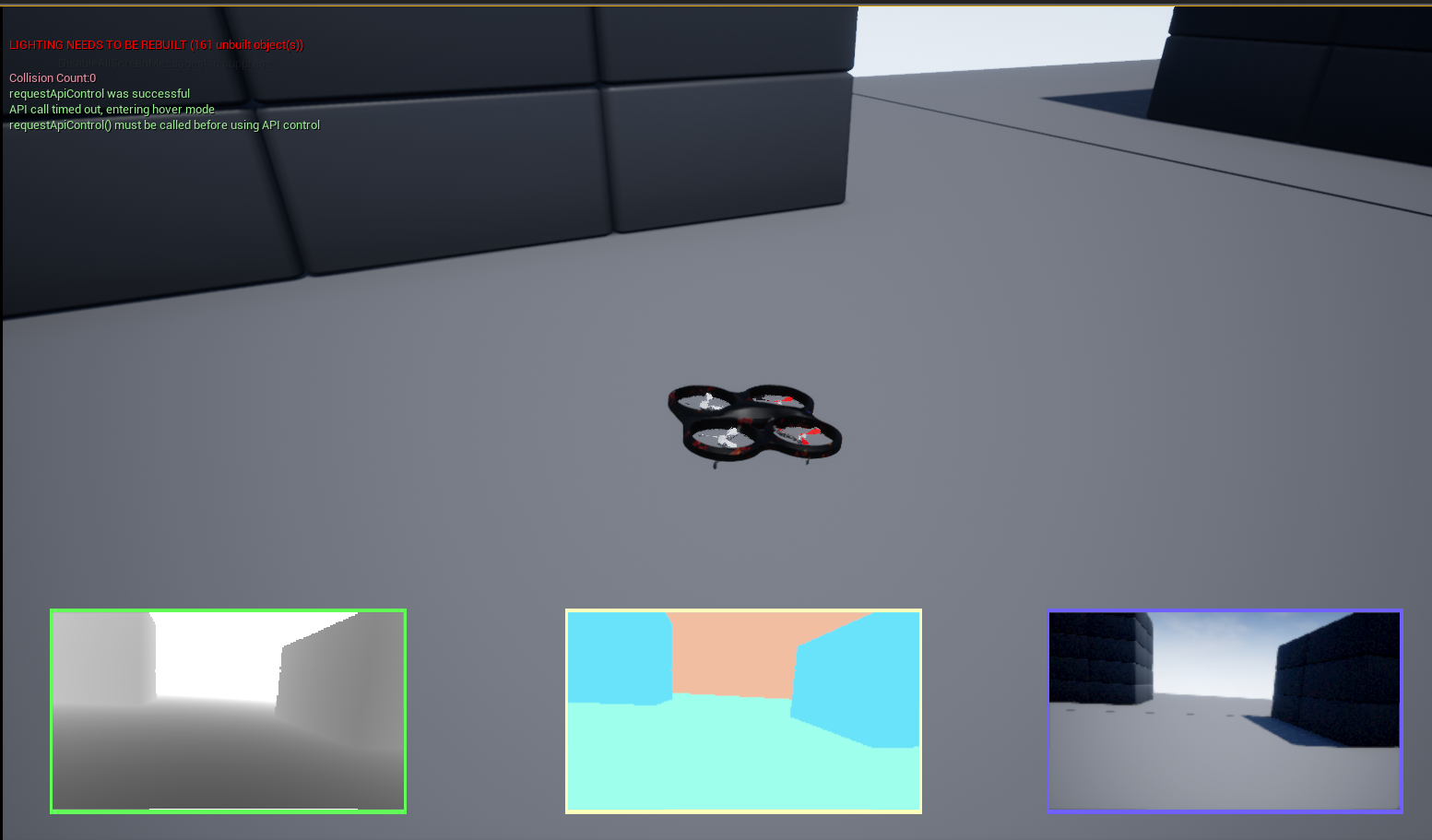}
\caption{Blocks Training Arena}
\end{figure}

\begin{figure}[!h]
\centering
\includegraphics[width=8cm]{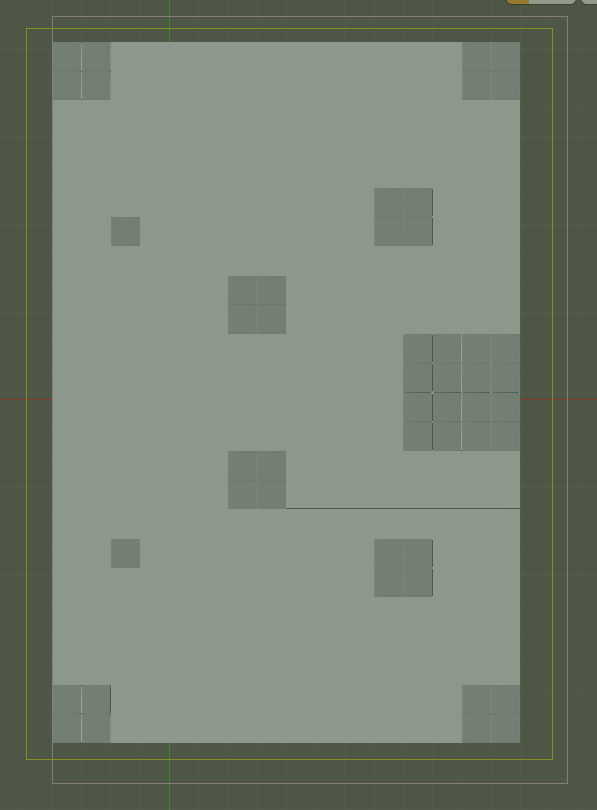}
\caption{Blocks Training Arena Top view}
\end{figure}

\begin{figure}[!h]
\centering
\includegraphics[width=8cm]{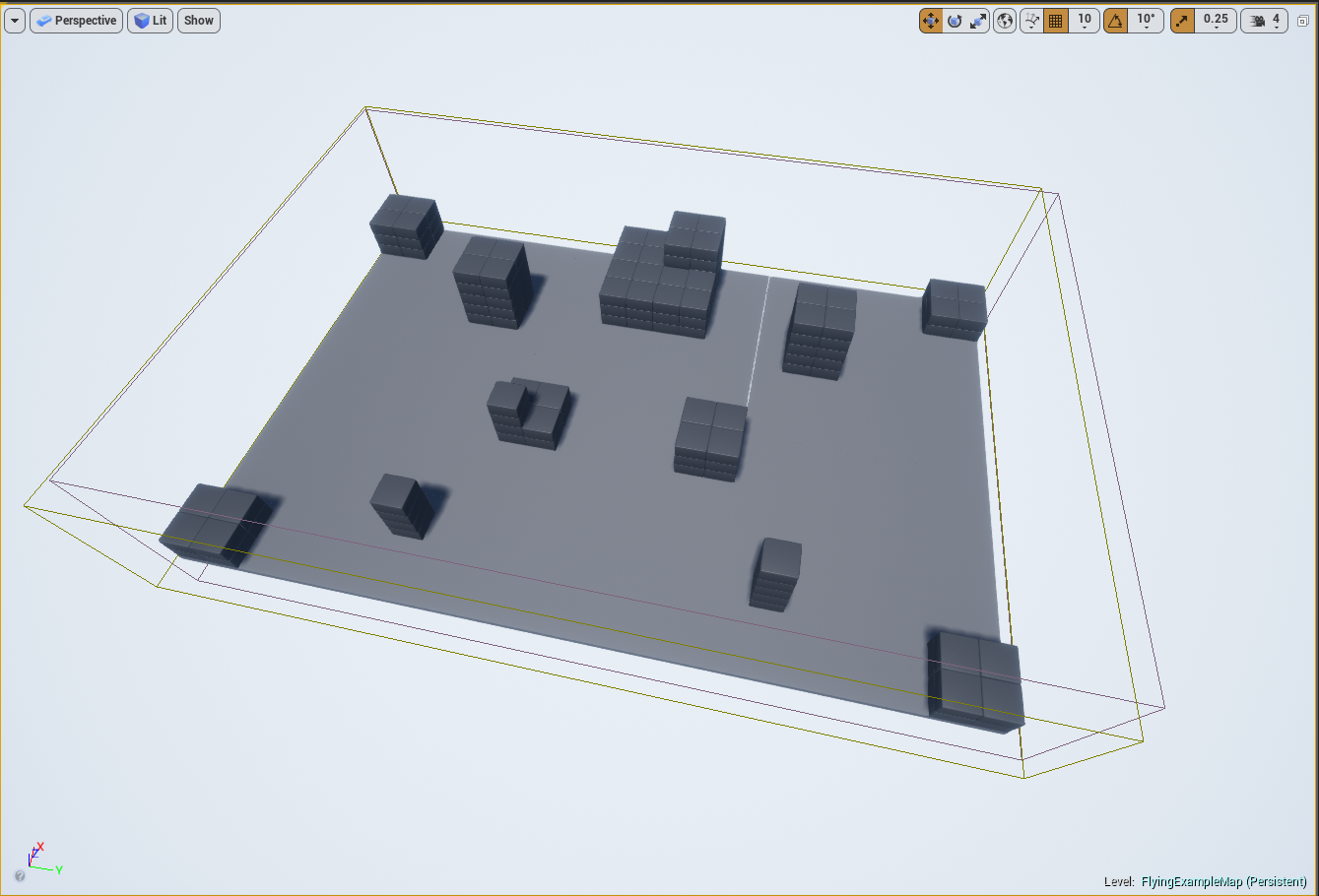}
\caption{Blocks Training Arena Isometric view}
\end{figure}

\subsection{Wobbles Training arena}

The Wobbles arena has multiple isolated training grounds
designed to train the quadcopter on different tasks. The Zone
A trains the agent to avoid cylindrical obstacles like pillars,
lamp posts, etc. The Zone B trains the agent to maneuver
around short walls. The Zone C is meant for sharp turn training.
The last Zone D is designed to test the performance of the
initial evasion training and also train for highly congested
environments. The obstacles in all the zones can be moved
dynamically to train for dynamic obstacles.

Primitive Training: Initially, the agent is trained with the
most basic obstacles to learn the baseline policies for avoiding
collisions. The agent is placed at one end and is expected to go
around this obstacle (be it a short wall or a cylinder) to
successfully complete the task.

Testing and improving: The robustness of this training can
be tested by running the primitively trained DQN in Zone D.
Although the agent is not expected to successfully traverse this
zone, the primitive training actually acts as a good initialization
and bolsters faster convergence to learn traversing through
Zone D. It also helps in obstacle generalization reducing
possible overfitting in the DQN.

The Zone C is used to train the agent to recognize and
execute sharp turns. The agent hasn’t been tested rigorously in
this Zone but it is an integral part of our future plan.
The final phase of training is expected to include all the
zones by randomly assigning a zone to the agent during each
training epoch. The agent will thus learn to generalize its
policy as the encountered features would be in random order.

\begin{figure}[!h]
\centering
\includegraphics[width=8cm]{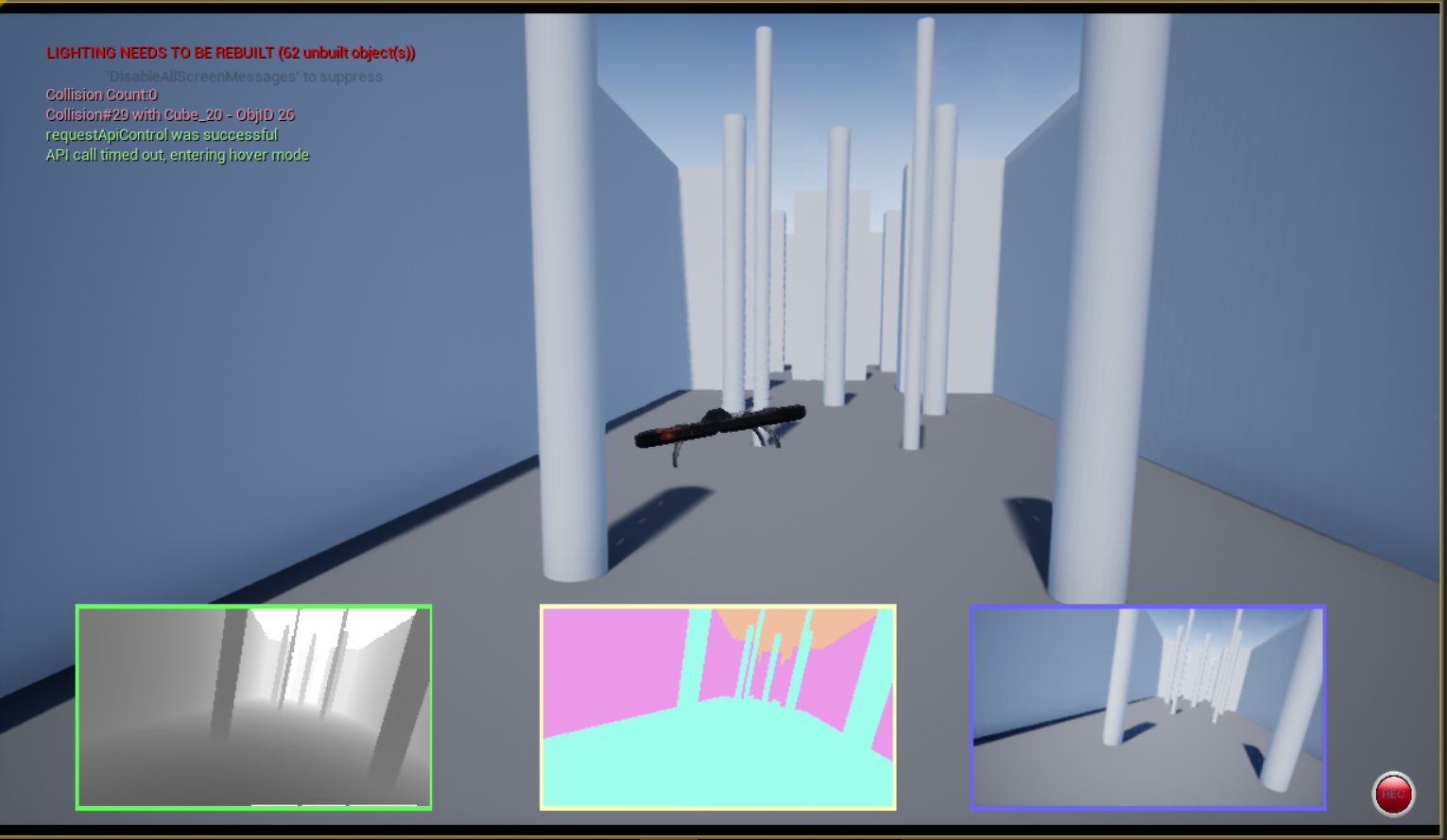}
\caption{Wobbles Training Arena}
\end{figure}

\begin{figure}[!h]
\centering
\includegraphics[width=8cm]{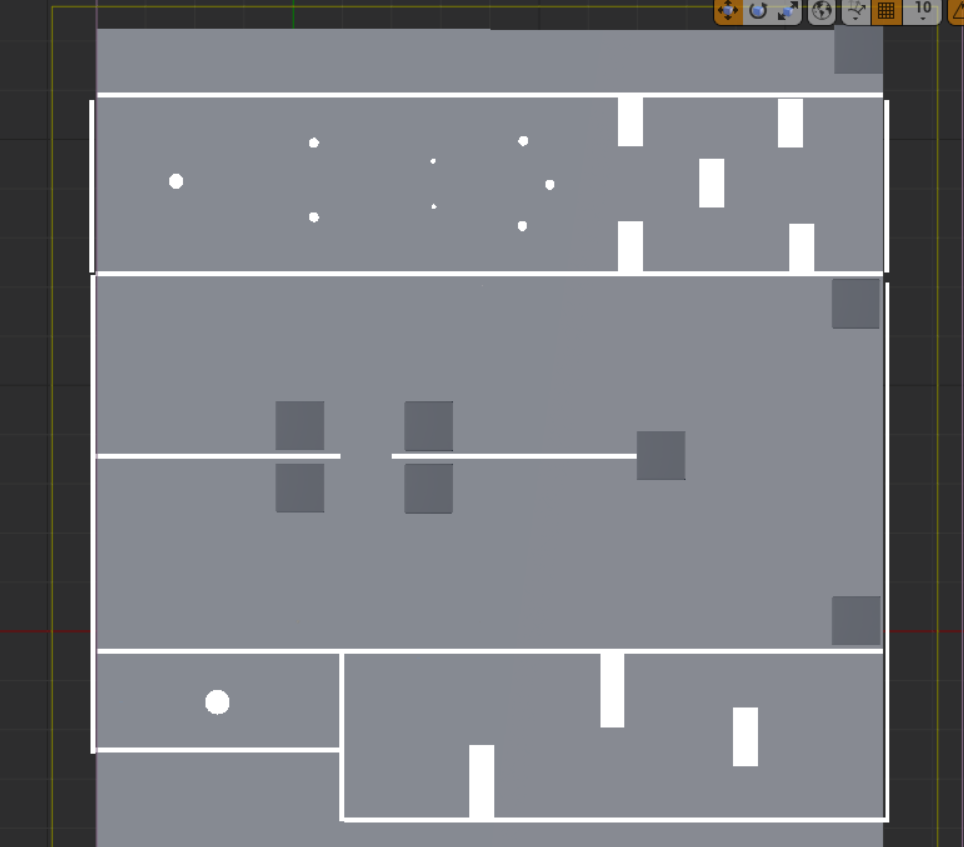}
\caption{Woobles Training Arena Top view}
\end{figure}

\begin{figure}[!h]
\centering
\includegraphics[width=8cm]{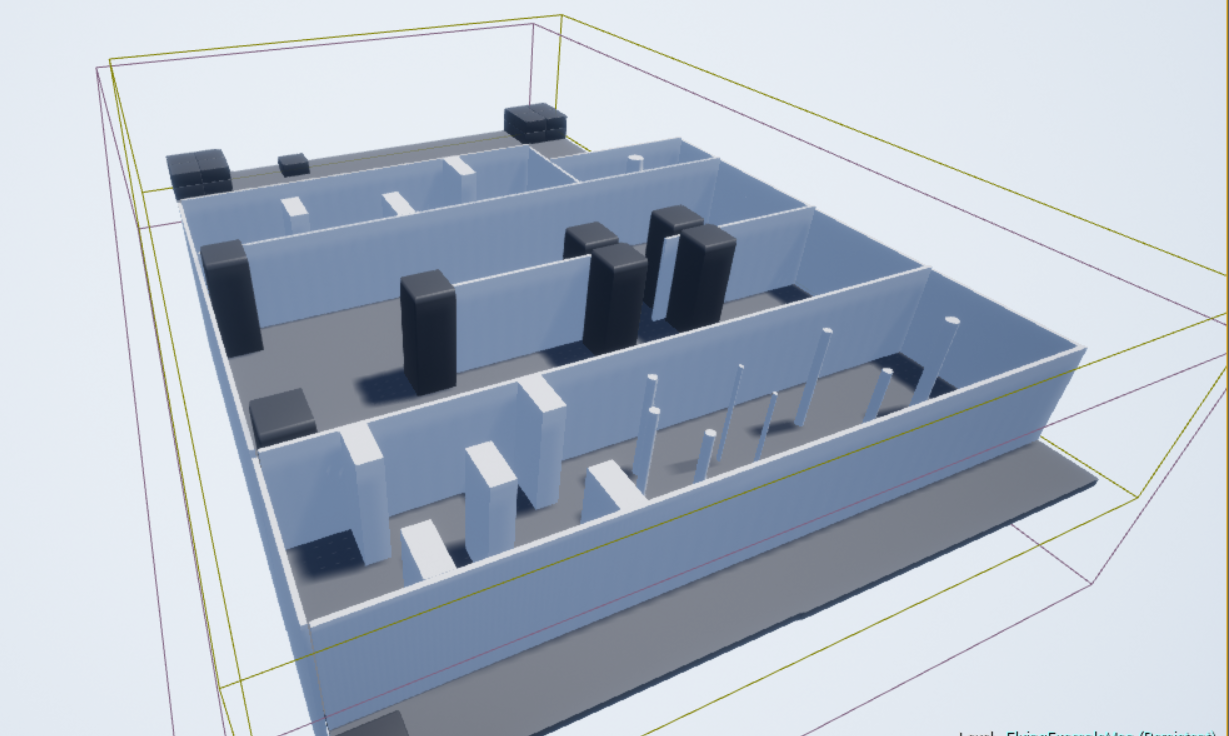}
\caption{Woobles Training Arena Isometric view}
\end{figure}

\section{RESULTS}
\subsection{Blocks environment trials}

In the blocks environment we found that the main
bottleneck for the DQN network to successfully recognize an
obstacle came from the field of view of the front centered
camera. Initially if the quadcopter is spawned very close to a
block, we found that it occluded the entire perspective of the
camera frame. In such cases the network had very less information
for choosing an action which would successfully avoid obstacles in front.

In the early experiments the reward was varied only based
upon the distance to the goal points and collisions. However,
we noticed that when the quadcopter randomly executed
opposite steering angles in a fast succession it would cause the
quadcopter and the camera frame to wobble which might
confuse the network. So we introduced a negative reward on
sharper yaw rates and we observed an improvement in
navigation and decrease in episode lengths.

As one can see, in the initial episodes, the quadcopter agent
is executing sharp turns which results in the roll angle to range
more than +- 30 degrees. After a few iterations, it successfully
learns to associates the negatively reward as the penalty for
such actions. A similar trend can be observed in the pitch
angles where the quad is holding a negative pitch angle so as
to maintain its forward velocity. As the training continued, we
observed that the agent successfully learned to stabilize itself,
reducing the number of sharp action inputs and executes
smooth turns.

\begin{figure}[!h]
\centering
\includegraphics[width=8cm]{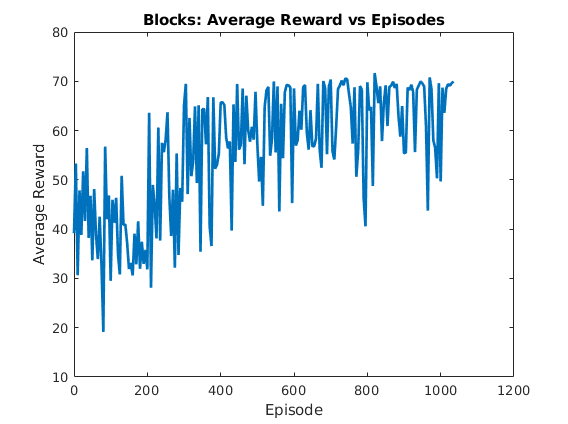}
\caption{Blocks Arena Average Reward}
\end{figure}
   
\begin{figure}[!h]
\centering
\includegraphics[width=8cm]{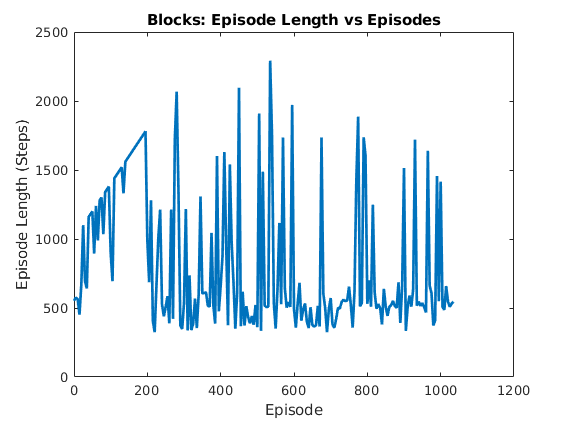}
\caption{Episode length in Blocks Arena}
\end{figure}

\begin{figure}[!h]
\centering
\includegraphics[width=8cm]{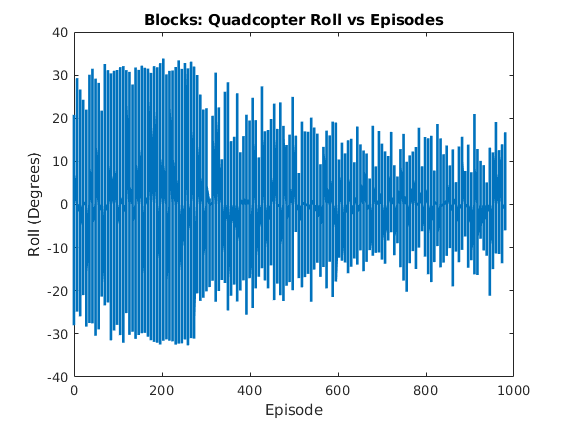}
\caption{Roll angle vs Episode for Blocks Arena}
\end{figure}

\begin{figure}[!h]
\centering
\includegraphics[width=8cm]{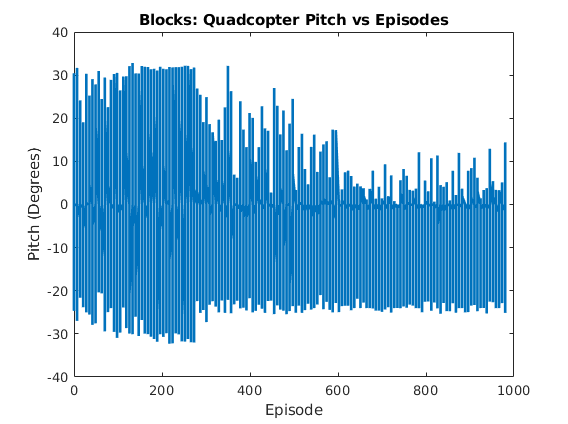}
\caption{Pitch angle vs Episode for Blocks Arena}
\end{figure}

\begin{figure}[!h]
\centering
\includegraphics[width=8cm]{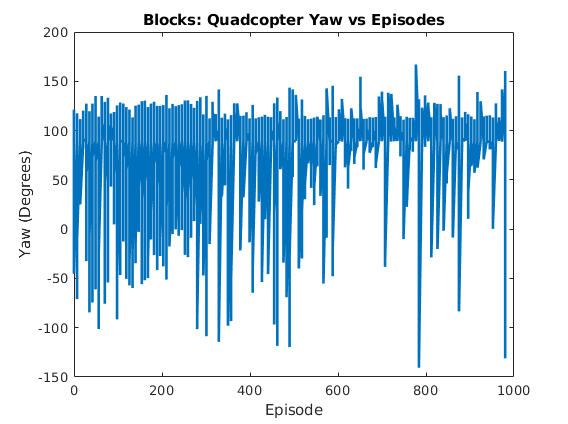}
\caption{Yaw angle vs Episode for Blocks Arena}
\end{figure}

\subsection{Wobbles Arena Zone D Trials:}

Wobble course was meant to simulate congested situations
that a quadcopter might encounter while navigating in urban
environments. It has a mix of cylindrical and short wall
obstacles and the agent is expected to distinguish them and
learn to fly around them or avoid them. The reward system in
Zone D is different to the blocks approach, here the route is
divided into checkpoints and the quadcopter is rewarded on
reaching the checkpoints while avoiding the obstacles.

The agent is initially trained in Zone A and B as a pre-
training steps to differentiate between walls and cylindrical
columns. The agent had to learn how to differentiate between
near and distant obstacles and also learn to avoid now a
combination of obstacles. We then started to train in the Zone
D. During the training course, it was observed that the agent
was trying to learn various local optimal policies and morphing
them upon failure. By a stochastic policy, we made sure it
doesn’t follow the same path while training, as sometimes it
would complete the Zone with a suboptimal policy. This is
essential especially in terms of learning to recover from fatal
states. After 1000 episodes of training, the quadcopter was able
to navigate sub-optimally through the zones. However, it used
to crash with the surrounding walls in intermediate episodes at
times. We suspect that this might be a result of uniform
gradient of the wall observed in the depth perspective image
which confuses the obstacle detection network.

We also see that the episode reward is directly related to
length of the episode. Thus, the agent successfully tries to
avoid obstacles while moving close to goal in the long running
episodes.

\begin{figure}[!h]
\centering
\includegraphics[width=8cm]{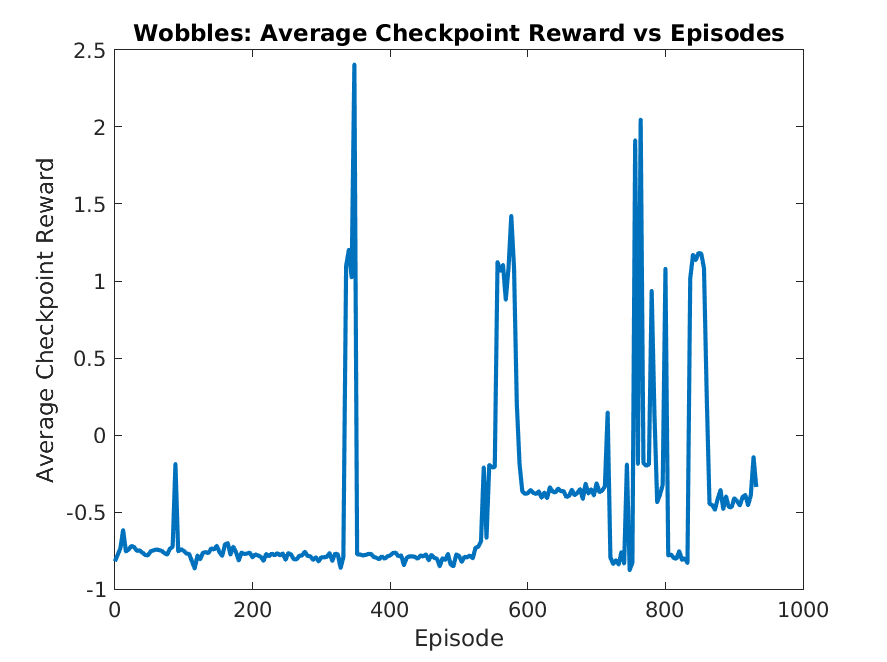}
\caption{Woobles Arena Average Reward}
\end{figure}
   
\begin{figure}[!h]
\centering
\includegraphics[width=8cm]{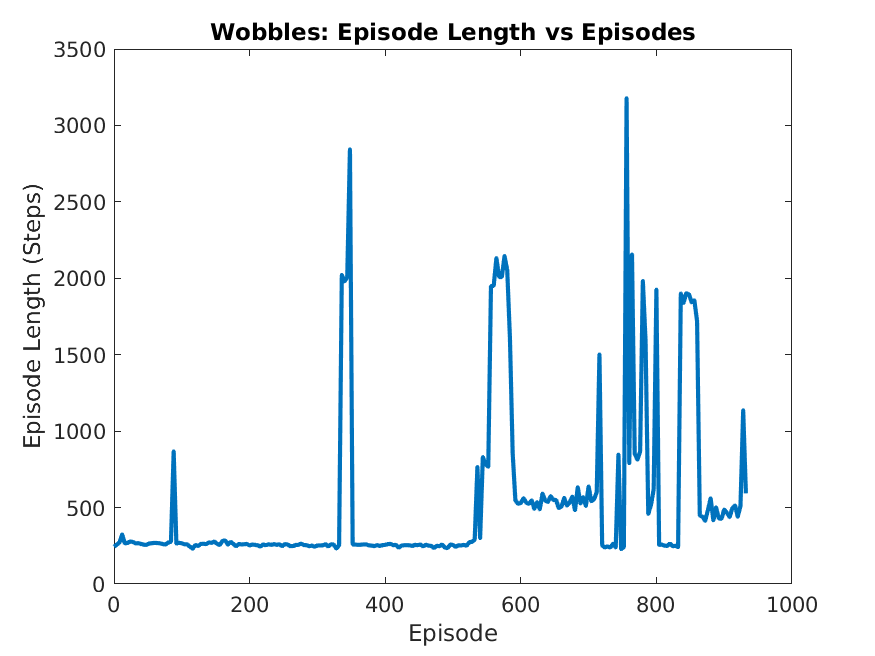}
\caption{Episode length in Woobles Arena}
\end{figure}

\section{FUTURE WORK}

The directional coordinate error can be merged at the
hidden layers, to let the network get a better idea of its position
in the environment with the environment visual information.
We can also use left and right camera to get a single
concatenated surrounding image and train the network based
on that, to get wider angle observations of the surrounding [9].
This method can be also extended to 360 degree cameras with
photometric error correction. Also we can experiment with
Dueling DQN which might solve the issue of obstacle
detection present at a far/near distance.

\section{CHALLENGES FACED}
\subsection{Modelling real-life complexity with simpler rewards}
It was to decide the state and action space for the navigation
problem. We initially decided to tackle it as a gridworld
problem by discretizing the state and action spaces in their
domains but found perception and localization to be a
challenge in the simulation. We finally decided with using
function approximation on visual cues and virtual GPS in
unison. Our action space was discretized to use 3 values of roll
for Wobbles arena and 5 values of yaw-rate for Blocks Arena.

The reward function had to be designed considering the
subgoals in mind as well as simultaneously keeping it simple
enough to avoid local optimal policies.

\subsection{Software challenges faced}
AirSim, being relatively new simulator, had to be studied
at an API level to understand the way in which different motion
primitives were implemented to be able to define our action
space clearly.

As we conducted the training sessions, we observed that the
Simulator would freeze randomly. We diagnosed the cause of
the problem to be Remote Procedural Calls timing out due to
unknown thread delays. As a workaround, we implemented
functions to save the state of the model and parameters and load
this data when running the training again.

The predefined controllers for motion with fixed yaw were
observed drift over time. We had to implement a secondary
controller correcting this drift at every step to keep the heading
direction of the quadcopter constant.

Since the predefined environments provided were limited
and none of them could be used for intensive training, we
developed our own ‘Wobbles’ training arena for learning
collision free maneuvers.

\section{CONCLUSIONS}
This paper presents an implementation of Double Deep Q
Learning to make the quadcopter with a depth camera learn an
acceptable policy to avoid obstacles. The model is trained in a
custom training arena containing different types of obstacles.
The results do show a gradual improvement in the policy as the
training proceeds. However, a large number of training will be
needed to generalize the obstacle avoidance skills. Locally
optimal policies learnt during the training course does show
that collision free navigation is possible solely using visual
cues.

This work is just a step in the direction of camera assisted
fully autonomous navigation using quadcopters. Further
improvement can be done by adding target displacement as a
part of the state. Enhancements to the current DDQN
framework like Dueling Networks can help in faster
convergence of policies.
\addtolength{\textheight}{-12cm}   








\end{document}